\RequirePackage[hyphens]{url}
\documentclass{article} 
\pdfoutput=1
\usepackage{iclr2017_conference,times}
\usepackage{hyperref}
\usepackage{url}
\usepackage[pdftex]{graphicx}
\usepackage[font=footnotesize]{caption}
\usepackage[utf8]{inputenc} 

\title{An Analysis of Deep Neural Network Models for Practical Applications}

\author{
  Alfredo Canziani \& Eugenio Culurciello\\
  Weldon School of Biomedical Engineering\\
  Purdue University\\
  \texttt{\{canziani,euge\}@purdue.edu}
  \And
  Adam Paszke\\
  Faculty of Mathematics, Informatics and Mechanics\\
  University of Warsaw\\
  \texttt{a.paszke@students.mimuw.edu.pl}\\
}

\begin{document}

\maketitle

\begin{abstract}
Since the emergence of Deep Neural Networks (DNNs) as a prominent technique in the field of computer vision, the ImageNet classification challenge has played a major role in advancing the state-of-the-art.
While accuracy figures have steadily increased, the resource utilisation of winning models has not been properly taken into account.
In this work, we present a comprehensive analysis of important metrics in practical applications: accuracy, memory footprint, parameters, operations count, inference time and power consumption.
Key findings are: (1) power consumption is independent of batch size and architecture; (2) accuracy and inference time are in a hyperbolic relationship; (3) energy constraint is an upper bound on the maximum achievable accuracy and model complexity; (4) the number of operations is a reliable estimate of the inference time.
We believe our analysis provides a compelling set of information that helps design and engineer efficient DNNs.
\end{abstract}

\section{Introduction}
Since the breakthrough in 2012 ImageNet competition \citep{russakovsky2015imagenet} achieved by AlexNet \citep{krizhevsky2012imagenet} --- the first entry that used a Deep Neural Network (DNN) --- several other DNNs with increasing complexity have been submitted to the challenge in order to achieve better performance.

In the ImageNet classification challenge, the ultimate goal is to obtain the highest accuracy in a multi-class classification problem framework, regardless of the actual inference time.
We believe that this has given rise to several problems. 
Firstly, it is now normal practice to run several trained instances of a given model over multiple similar instances of each validation image.
This practice, also know as model averaging or ensemble of DNNs, dramatically increases the amount of computation required at inference time to achieve the published accuracy.
Secondly, model selection is hindered by the fact that different submissions are evaluating their (ensemble of) models a different number of times on the validation images, and therefore the reported accuracy is biased on the specific sampling technique (and ensemble size).
Thirdly, there is currently no incentive in speeding up inference time, which is a key element in practical applications of these models, and affects resource utilisation, power-consumption, and latency.

This article aims to compare state-of-the-art DNN architectures, submitted for the ImageNet challenge over the last 4 years, in terms of computational requirements and accuracy.
We compare these architectures on multiple metrics related to resource utilisation in actual deployments: accuracy, memory footprint, parameters, operations count, inference time and power consumption.
The purpose of this paper is to stress the importance of these figures, which are essential hard constraints for the optimisation of these networks in practical deployments and applications.

\begin{figure}[!t]
  \centering
  \includegraphics[width=1.0\textwidth]{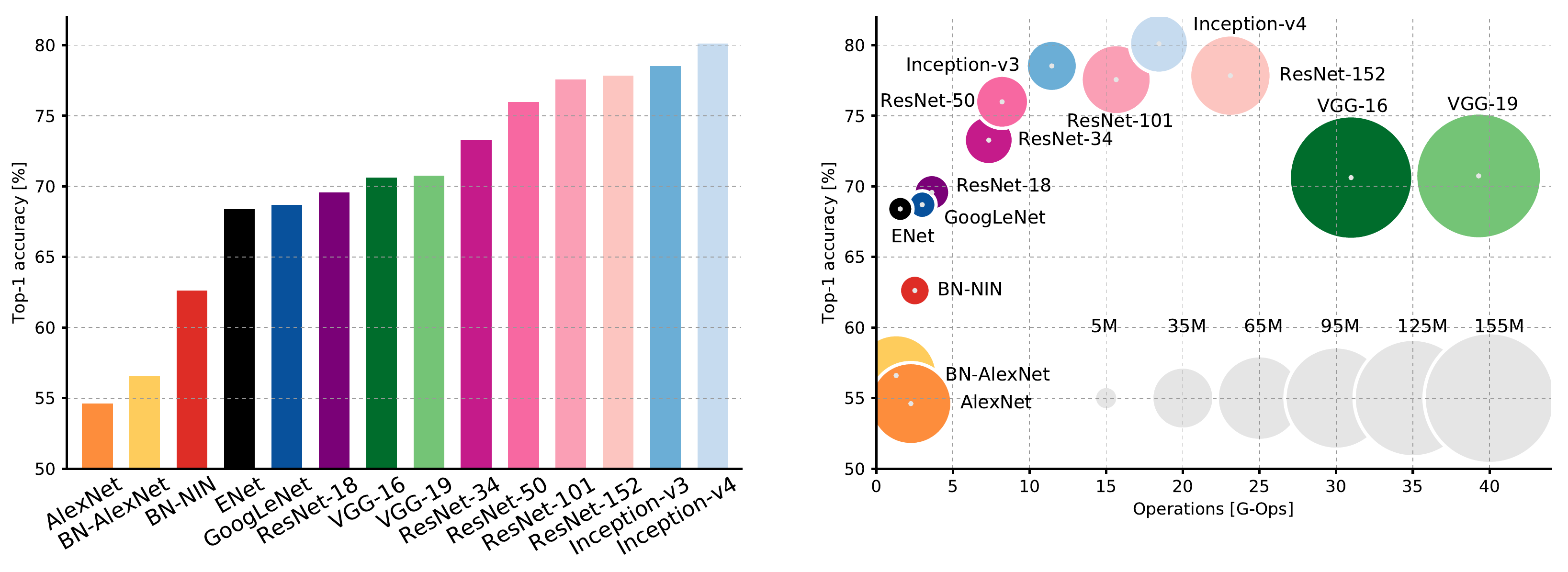}
  \begin{minipage}[t]{.48\textwidth}
    \centering
    \caption{
      \textbf{Top1 \emph{vs.}\ network.}
      Single-crop top-1 validation accuracies for top scoring single-model architectures.
      We introduce with this chart our choice of colour scheme, which will be used throughout this publication to distinguish effectively different architectures and their correspondent authors.
Notice that networks of the same group share the same hue, for example ResNet are all variations of pink.
    }
    \label{fig:top1_vs_net}
  \end{minipage} \quad
  \begin{minipage}[t]{.48\textwidth}
    \centering
    \caption{
      \textbf{Top1 \emph{vs.}\ operations, size $\propto$ parameters.}
      Top-1 one-crop accuracy versus amount of operations required for a single forward pass.
      The size of the blobs is proportional to the number of network parameters; a legend is reported in the bottom right corner, spanning from $5 \times 10^6$ to $155 \times 10^6$ params.
      Both these figures share the same \emph{y}-axis, and the grey dots highlight the centre of the blobs.
    }
    \label{fig:top1_vs_ops}
  \end{minipage}
\end{figure}

\section{Methods} \label{s:method}
In order to compare the quality of different models, we collected and analysed the accuracy values reported in the literature.
We immediately found that different sampling techniques do not allow for a direct comparison of resource utilisation.
For example, central-crop (top-5 validation) errors of a single run of VGG-16\footnote{
  In the original paper this network is called VGG-D, which is the best performing network.
  Here we prefer to highlight the number of layer utilised, so we will call it VGG-16 in this publication.
} \citep{simonyan2014very} and GoogLeNet \citep{szegedy2014going} are $8.70\%$ and $10.07\%$ respectively, revealing that VGG-16 performs better than GoogLeNet.
When models are run with 10-crop sampling,\footnote{
  From a given image multiple patches are extracted: four corners plus central crop and their horizontal mirrored twins.
} then the errors become $9.33\%$ and $9.15\%$ respectively, and therefore VGG-16 will perform worse than GoogLeNet, using a single central-crop.
For this reason, we decided to base our analysis on re-evaluations of top-1 accuracies\footnote{
  Accuracy and error rate always sum to $100$, therefore in this paper they are used interchangeably.
} for all networks with a single central-crop sampling technique \citep{zagoruyko2016imagenet}.

For inference time and memory usage measurements we have used Torch7 \citep{collobert2011torch7} with cuDNN-v5 \citep{chetlur2014cudnn} and CUDA-v8 back-end.
All experiments were conducted on a JetPack-2.3 NVIDIA Jetson TX1 board \citep{nvidia2015jetson}: an embedded visual computing system with a 64-bit ARM® A57 CPU, a 1 T-Flop/s 256-core NVIDIA Maxwell GPU and 4 GB LPDDR4 of shared RAM.
We use this resource-limited device to better underline the differences between network architecture, but similar results can be obtained on most recent GPUs, such as the NVIDIA K40 or Titan X, to name a few.
Operation counts were obtained using an open-source tool that we developed \citep{paszke2016opcounter}.
For measuring the power consumption, a Keysight 1146B Hall effect current probe has been used with a Keysight MSO-X 2024A $200\,\mathrm{MHz}$ digital oscilloscope with a sampling period of $2\,\mathrm{s}$ and $50\,\mathrm{kSa/s}$ sample rate.
The system was powered by a Keysight E3645A GPIB controlled DC power supply.

\section{Results}

In this section we report our results and comparisons.
We analysed the following DDNs: AlexNet \citep{krizhevsky2012imagenet}, batch normalised AlexNet \citep{zagoruyko2016imagenet}, batch normalised Network In Network (NIN) \citep{lin2013network}, ENet \citep{paszke2016enet} for ImageNet \citep{culurciello2016training}, GoogLeNet \citep{szegedy2014going}, VGG-16 and -19 \citep{simonyan2014very}, \mbox{ResNet-18}, \mbox{-34}, \mbox{-50}, \mbox{-101} and \mbox{-152} \citep{he2015deep}, Inception-v3 \citep{szegedy2015rethinking} and Inception-v4 \citep{szegedy2016inception} since they obtained the highest performance, in these four years, on the ImageNet \citep{russakovsky2015imagenet} challenge.

\subsection{Accuracy} \label{ss:acc}

Figure \ref{fig:top1_vs_net} shows one-crop accuracies of the most relevant entries submitted to the ImageNet challenge, from the AlexNet \citep{krizhevsky2012imagenet}, on the far left, to the best performing Inception-v4 \citep{szegedy2016inception}.
The newest ResNet and Inception architectures surpass all other architectures by a significant margin of at least $7\%$.

Figure \ref{fig:top1_vs_ops} provides a different, but more informative view of the accuracy values, because it also visualises computational cost and number of network's parameters.
The first thing that is very apparent is that VGG, even though it is widely used in many applications, is by far the most expensive architecture --- both in terms of computational requirements and number of parameters.
Its 16- and 19-layer implementations are in fact isolated from all other networks.
The other architectures form a steep straight line, that seems to start to flatten with the latest incarnations of Inception and ResNet.
This might suggest that models are reaching an inflection point on this data set.
At this inflection point, the costs --- in terms of complexity --- start to outweigh gains in accuracy.
We will later show that this trend is hyperbolic.

\subsection{Inference Time}

\begin{figure}[!t]
  \centering
  \includegraphics[width=1.0\textwidth]{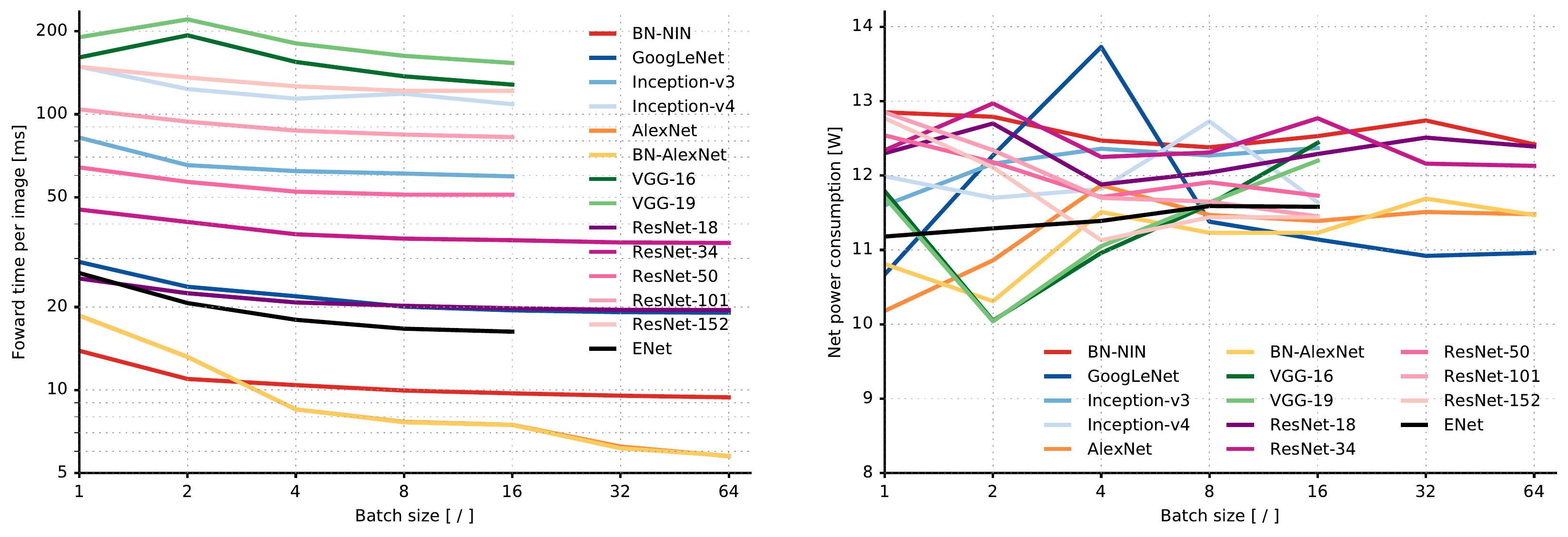}
  \begin{minipage}[t]{.48\textwidth}
    \centering
    \caption{
      \textbf{Inference time \emph{vs.}\ batch size.}
      This chart show inference time across different batch sizes with a logarithmic ordinate and logarithmic abscissa.
      Missing data points are due to lack of enough system memory required to process larger batches.
      A speed up of $3\times$ is achieved by AlexNet due to better optimisation of its fully connected layers for larger batches.
    }
    \label{fig:time_vs_batch}
  \end{minipage} \quad
  \begin{minipage}[t]{.48\textwidth}
    \centering
    \caption{
      \textbf{Power \emph{vs.}\ batch size.}
      Net power consumption (due only to the forward processing of several DNNs) for different batch sizes.
      The idle power of the TX1 board, with no HDMI screen connected, was $1.30\,\mathrm{W}$ on average.
      The max frequency component of power supply current was $1.4\,\mathrm{kHz}$, corresponding to a Nyquist sampling frequency of $2.8\,\mathrm{kHz}$.
    }
    \label{fig:power_vs_batch}
  \end{minipage}
\end{figure}

Figure \ref{fig:time_vs_batch} reports inference time per image on each architecture, as a function of image batch size (from 1 to 64).
We notice that VGG processes one image in a fifth of a second, making it a less likely contender in real-time applications on an NVIDIA TX1.
AlexNet shows a speed up of roughly $3\times$ going from batch of 1 to 64 images, due to weak optimisation of its fully connected layers.
It is a very surprising finding, that will be further discussed in the next subsection.

\subsection{Power}

Power measurements are complicated by the high frequency swings in current consumption, which required high sampling current read-out to avoid aliasing.
In this work, we used a $200\,\mathrm{MHz}$ digital oscilloscope with a current probe, as reported in section \ref{s:method}.
Other measuring instruments, such as an AC power strip with $2\,\mathrm{Hz}$ sampling rate, or a GPIB controlled DC power supply with $12\,\mathrm{Hz}$ sampling rate, did not provide enough bandwidth to properly conduct power measurements.

In figure \ref{fig:power_vs_batch} we see that the power consumption is mostly independent with the batch size.
Low power values for AlexNet (batch of 1) and VGG (batch of 2) are associated to slower forward times per image, as shown in figure \ref{fig:time_vs_batch}.

\subsection{Memory}

\begin{figure}[!t]
  \centering
  \includegraphics[width=1.0\textwidth]{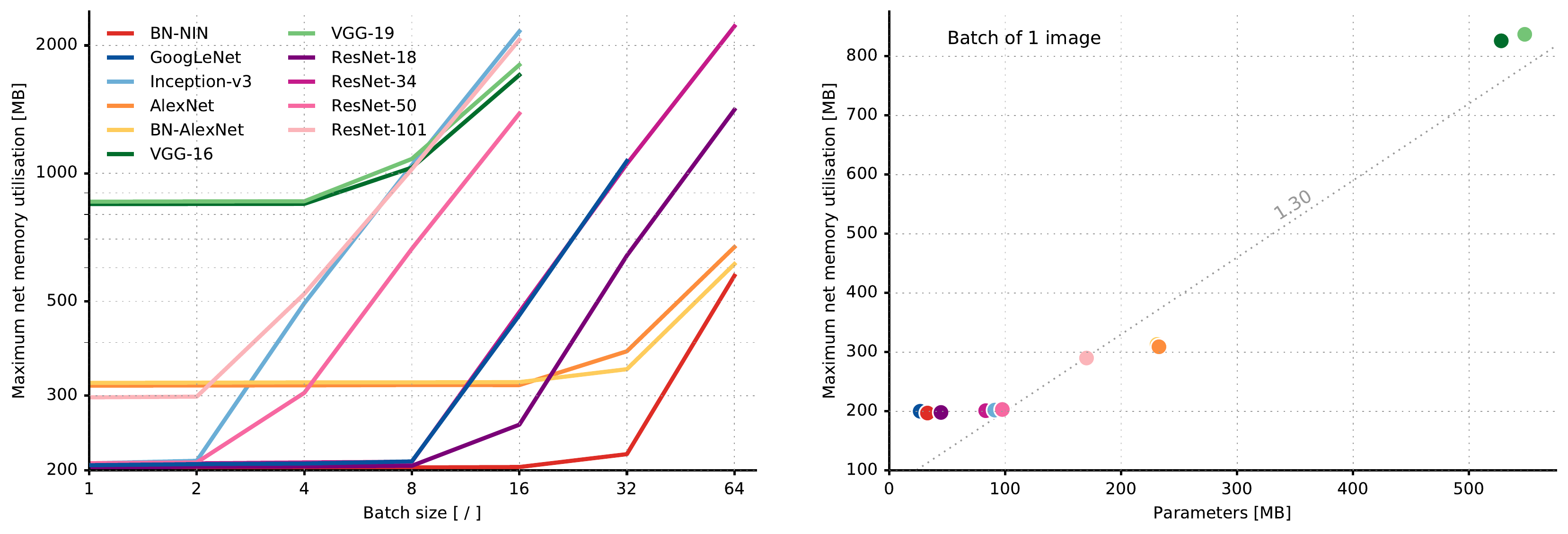}
  \begin{minipage}[t]{.48\textwidth}
    \centering
    \caption{
      \textbf{Memory \emph{vs.}\ batch size.}
      Maximum system memory utilisation for batches of different sizes.
      Memory usage shows a knee graph, due to the network model memory static allocation and the variable memory used by batch size.
    }
    \label{fig:mem_vs_batch}
  \end{minipage} \quad
  \begin{minipage}[t]{.48\textwidth}
    \centering
    \caption{
      \textbf{Memory \emph{vs.}\ parameters count.}
      Detailed view on static parameters allocation and corresponding memory utilisation.
      Minimum memory of $200\,\mathrm{MB}$, linear afterwards with slope $1.30$.
    }
    \label{fig:mem_vs_par}
  \end{minipage}
\end{figure}

We analysed system memory consumption of the TX1 device, which uses shared memory for both CPU and GPU.
Figure \ref{fig:mem_vs_batch} shows that the maximum system memory usage is initially constant and then raises with the batch size.
This is due the initial memory allocation of the network model --- which is the large static component --- and the contribution of the memory required while processing the batch, proportionally increasing with the number of images.
In figure \ref{fig:mem_vs_par} we can also notice that the initial allocation never drops below $200\,\mathrm{MB}$, for network sized below $100\,\mathrm{MB}$, and it is linear afterwards, with respect to the parameters and a slope of $1.30$.

\subsection{Operations}

\begin{figure}[!t]
  \centering
  \includegraphics[width=1.0\textwidth]{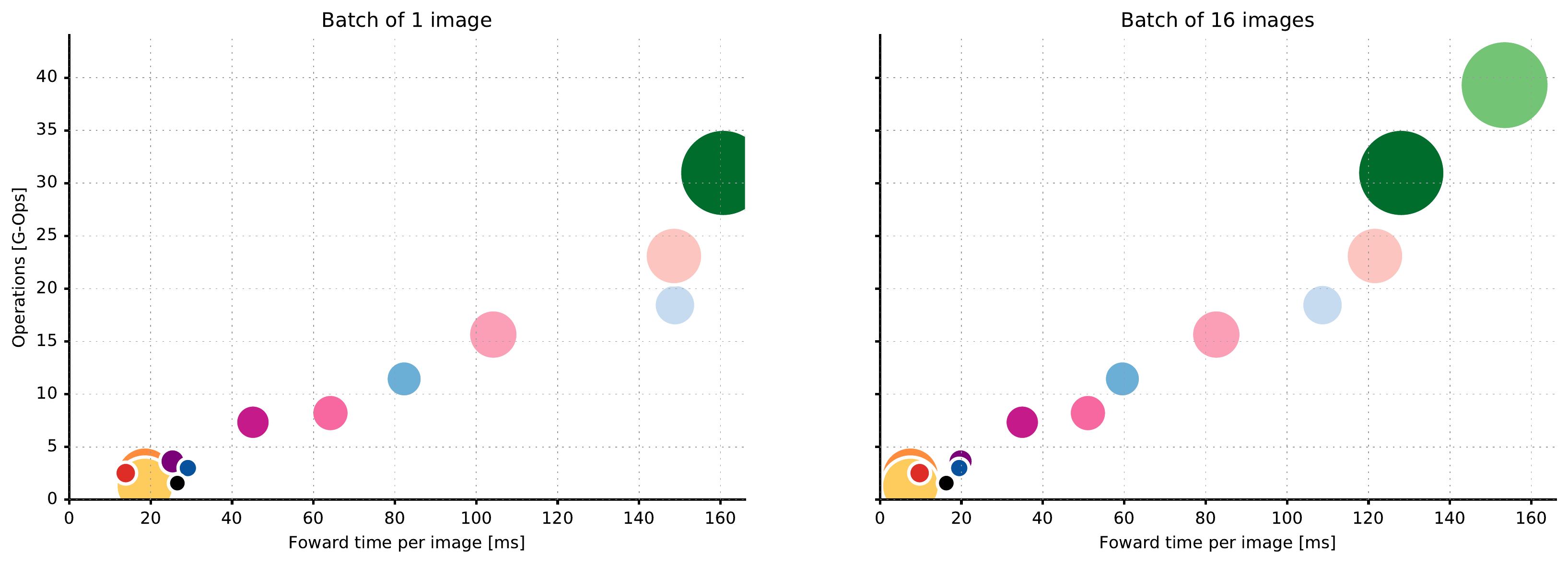}
  \caption{
    \textbf{Operations \emph{vs.}\ inference time, size $\propto$ parameters.}
    Relationship between operations and inference time, for batches of size 1 and 16 (biggest size for which all architectures can still run).
    Not surprisingly, we notice a linear trend, and therefore operations count represent a good estimation of inference time.
    Furthermore, we can notice an increase in the slope of the trend for larger batches, which correspond to shorter inference time due to batch processing optimisation.
  }
  \label{fig:ops_vs_time}
\end{figure}

Operations count is essential for establishing a rough estimate of inference time and hardware circuit size, in case of custom implementation of neural network accelerators.
In figure \ref{fig:ops_vs_time}, for a batch of 16 images, there is a linear relationship between operations count and inference time per image.
Therefore, at design time, we can pose a constraint on the number of operation to keep processing speed in a usable range for real-time applications or resource-limited deployments.

\subsection{Operations and Power} \label{ss:ops_and_power}

\begin{figure}[!t]
  \centering
  \includegraphics[width=1.0\textwidth]{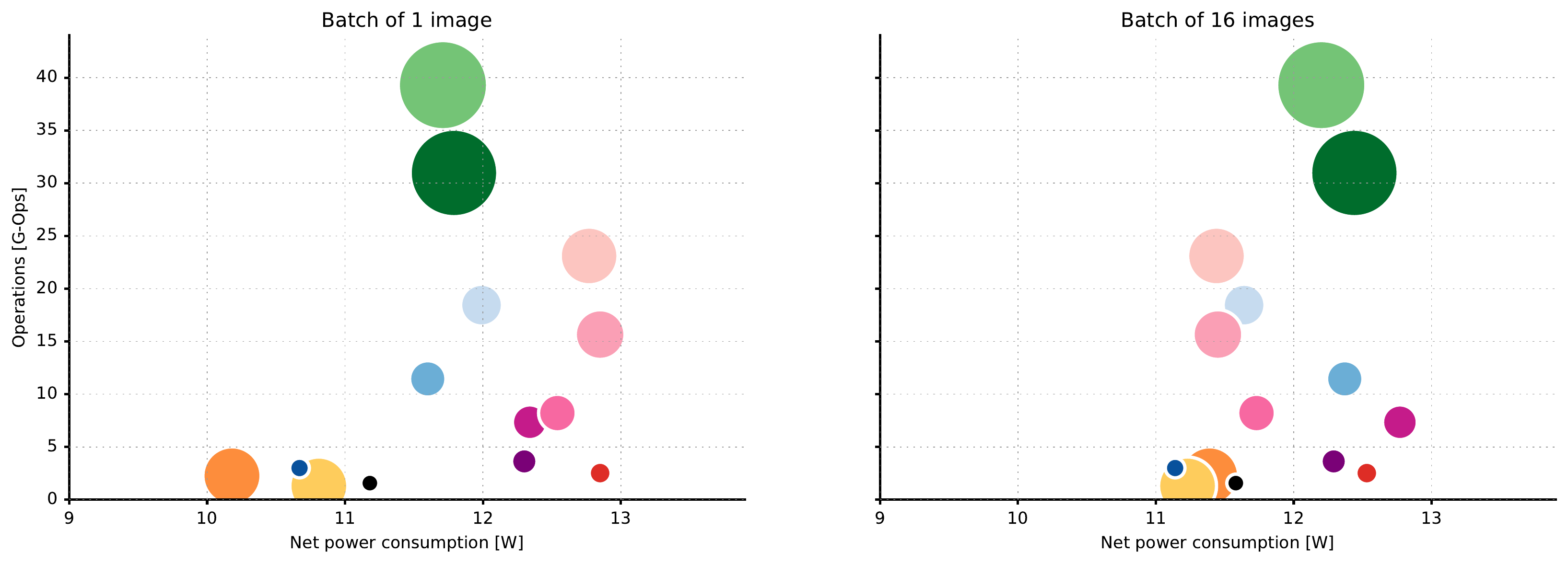}
  \caption{
    \textbf{Operations \emph{vs.}\ power consumption, size $\propto$ parameters.}
    Independency of power and operations is shown by a lack of directionality of the distributions shown in these scatter charts.
    Full resources utilisation and lower inference time for AlexNet architecture is reached with larger batches.
  }
  \label{fig:ops_vs_power}
\end{figure}

In this section we analyse the relationship between power consumption and number of operations required by a given model.
Figure \ref{fig:ops_vs_power} reports that there is no specific power footprint for different architectures.
When full resources utilisation is reached, generally with larger batch sizes, all networks consume roughly an additional $11.8\,\mathrm{W}$, with a standard deviation of $0.7\,\mathrm{W}$.
Idle power is $1.30\,\mathrm{W}$.
This corresponds to the maximum system power at full utilisation.
Therefore, if energy consumption is one of our concerns, for example for battery-powered devices, one can simply choose the slowest architecture which satisfies the application minimum requirements.

\subsection{Accuracy and Throughput}

\begin{figure}[!t]
  \centering
  \includegraphics[width=1.0\textwidth]{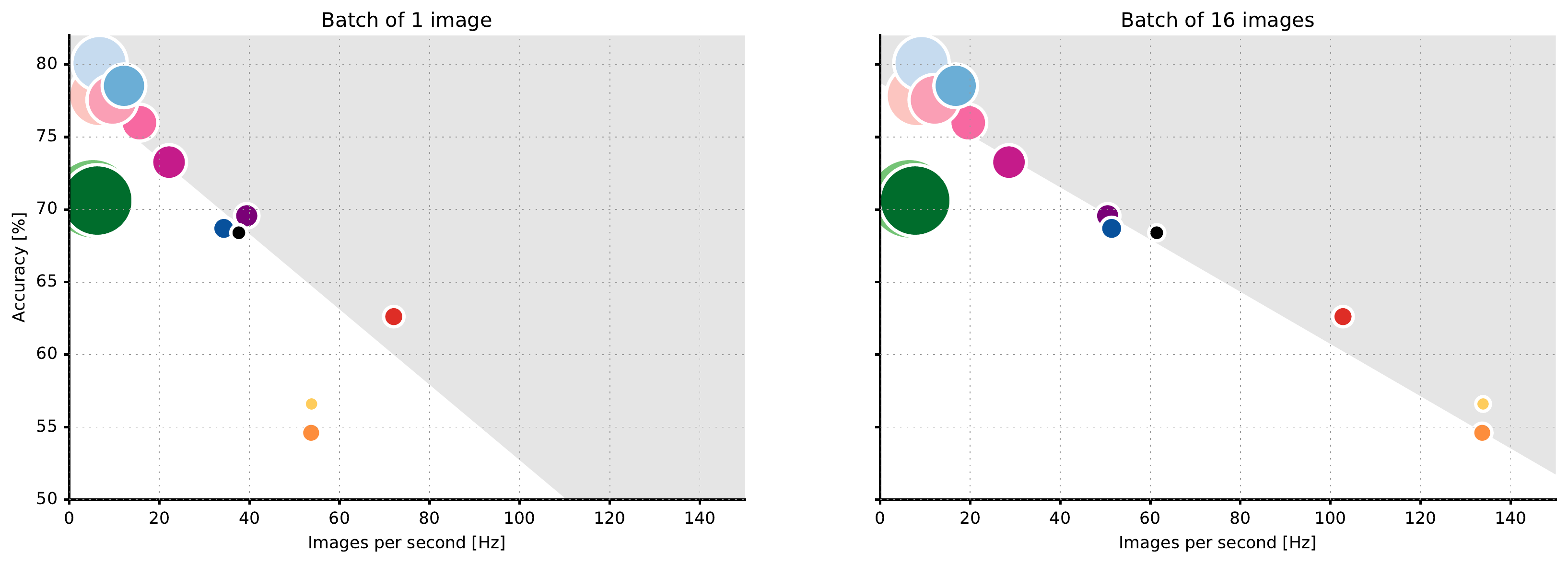}
  \caption{
    \textbf{Accuracy \emph{vs.}\ inferences per second, size $\propto$ operations.}
    Non trivial linear upper bound is shown in these scatter plots, illustrating the relationship between prediction accuracy and throughput of all examined architectures.
    These are the first charts in which the area of the blobs is proportional to the amount of operations, instead of the parameters count.
    We can notice that larger blobs are concentrated on the left side of the charts, in correspondence of low throughput, \emph{i.e.}\ longer inference times.
    Most of the architectures lay on the linear interface between the grey and white areas.
    If a network falls in the shaded area, it means it achieves exceptional accuracy or inference speed.
    The white area indicates a suboptimal region.
    \emph{E.g.}\ both AlexNet architectures improve processing speed as larger batches are adopted, gaining $80\,\mathrm{Hz}$.
  }
  \label{fig:acc_vs_fps}
\end{figure}

We note that there is a non-trivial linear upper bound between accuracy and number of inferences per unit time.
Figure \ref{fig:acc_vs_fps} illustrates that for a given frame rate, the maximum accuracy that can be achieved is linearly proportional to the frame rate itself.
All networks analysed here come from several publications, and have been independently trained by other research groups.
A linear fit of the accuracy shows all architecture trade accuracy \emph{vs}.\ speed.
Moreover, chosen a specific inference time, one can now come up with the theoretical accuracy upper bound when resources are fully utilised, as seen in section \ref{ss:ops_and_power}.
Since the power consumption is constant, we can even go one step further, and obtain an upper bound in accuracy even for an energetic constraint, which could possibly be an essential designing factor for a network that needs to run on an embedded system.

As the spoiler in section \ref{ss:acc} gave already away, the linear nature of the accuracy \emph{vs.}\ throughput relationship translates into a hyperbolical one when the forward inference time is considered instead.
Then, given that the operations count is linear with the inference time, we get that the accuracy has an hyperbolical dependency on the amount of computations that a network requires.

\subsection{Parameters Utilisation}

\begin{figure}[!t]
  \centering
  \includegraphics[width=0.625\textwidth]{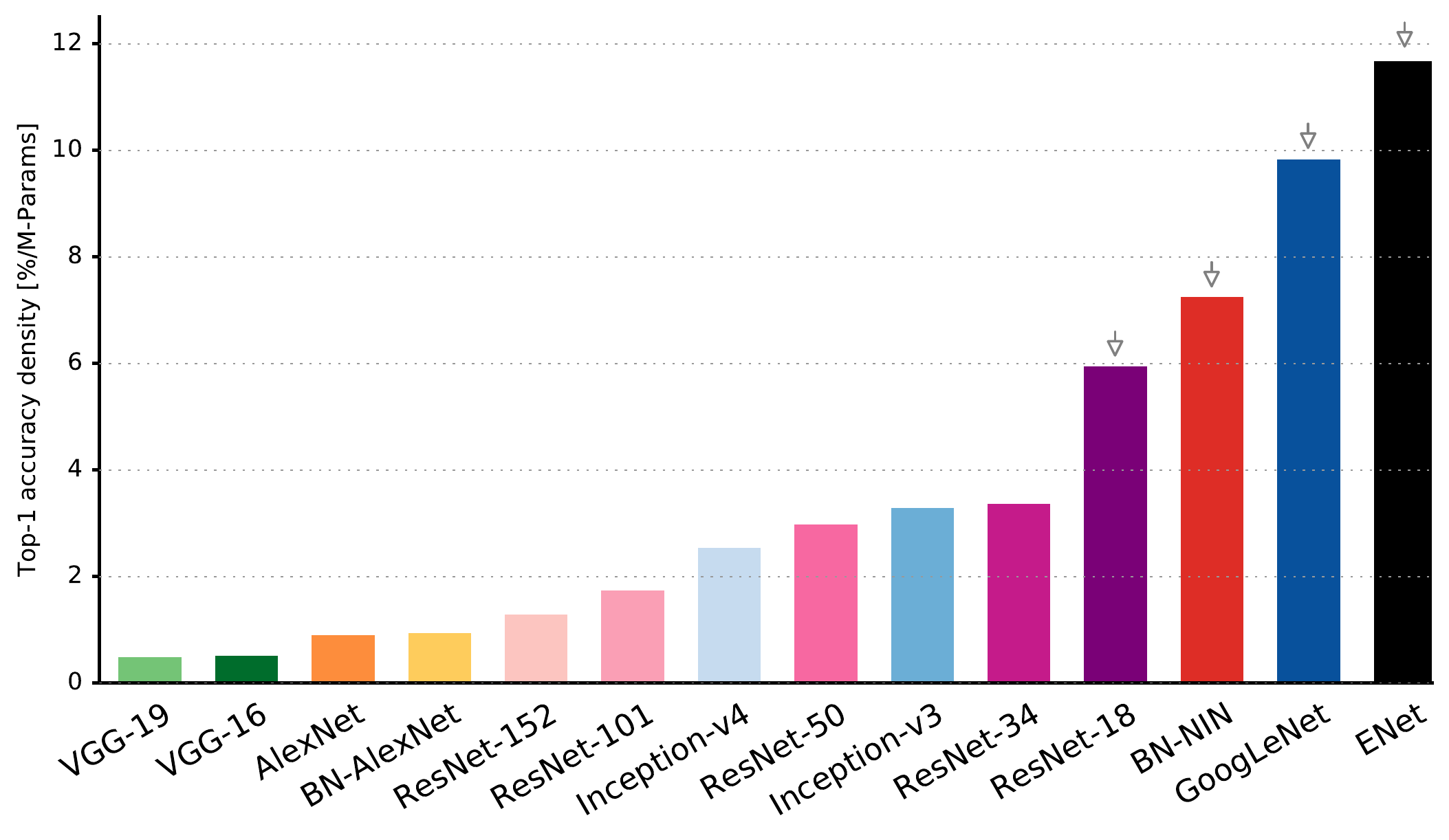}
  \caption{
    \textbf{Accuracy per parameter \emph{vs.}\ network.}
    Information density (accuracy per parameters) is an efficiency metric that highlight that capacity of a specific architecture to better utilise its parametric space.
    Models like VGG and AlexNet are clearly oversized, and do not take fully advantage of their potential learning ability.
    On the far right, ResNet-18, BN-NIN, GoogLeNet and ENet (marked by grey arrows) do a better job at ``squeezing'' all their neurons to learn the given task, and are the winners of this section.
  }
  \label{fig:acc_dens_vs_net}
\end{figure}

DNNs are known to be highly inefficient in utilising their full learning power (number of parameters / degrees of freedom).
Prominent work \citep{han2015deep} exploits this flaw to reduce network file size up to $50\times$, using weights pruning, quantisation and variable-length symbol encoding.
It is worth noticing that, using more efficient architectures to begin with may produce even more compact representations.
In figure \ref{fig:acc_dens_vs_net} we clearly see that, although VGG has a better accuracy than AlexNet (as shown by figure \ref{fig:top1_vs_net}), its information density is worse.
This means that the amount of degrees of freedom introduced in the VGG architecture bring a lesser improvement in terms of accuracy.
Moreover, ENet \citep{paszke2016enet} --- which we have specifically designed to be highly efficient and it has been adapted and retrained on ImageNet \citep{culurciello2016training} for this work --- achieves the highest score, showing that $24\times$ less parameters are sufficient to provide state-of-the-art results.

\section{Conclusions}

In this paper we analysed multiple state-of-the-art deep neural networks submitted to the ImageNet challenge, in terms of accuracy, memory footprint, parameters, operations count, inference time and power consumption.
Our goal is to provide insights into the design choices that can lead to efficient neural networks for practical application, and optimisation of the often-limited resources in actual deployments, which lead us to the creation of ENet --- or Efficient-Network --- for ImageNet.
We show that accuracy and inference time are in a hyperbolic relationship: a little increment in accuracy costs a lot of computational time.
We show that number of operations in a network model can effectively estimate inference time.
We show that an energy constraint will set a specific upper bound on the maximum achievable accuracy and model complexity, in terms of operations counts.
Finally, we show that ENet is the best architecture in terms of parameters space utilisation, squeezing up to $13\times$ more information per parameter used respect to the reference model AlexNet, and $24\times$ respect VGG-19.

\subsubsection*{Acknowledgments}

This paper would have not look so pretty without the \emph{Python Software Foundation}, the \texttt{mat\-plot\-lib} library and the communities of \emph{stackoverflow} and \TeX\ of \emph{StackExchange} which I ought to thank.
This work is partly supported by the Office of Naval Research (ONR) grants N00014-12-1-0167, N00014-15-1-2791 and MURI N00014-10-1-0278.
We gratefully acknowledge the support of NVIDIA Corporation with the donation of the TX1, Titan X, K40 GPUs used for this research.

{
  \footnotesize
  \selectfont
  \bibliography{ref}
  \bibliographystyle{iclr2017_conference}
}

\end{document}